\documentclass[lettersize,journal]{IEEEtran}
\ifCLASSINFOpdf
\else
\fi
\usepackage{xcolor}
\usepackage{array}
\usepackage{colortbl}
\definecolor{ceblue}{rgb}{ .608,  .761,  .902}
\definecolor{cellblue}{RGB}{102,  178,  255}
\usepackage{graphicx}
\usepackage{hyperref}
\usepackage{indentfirst}
\usepackage{multirow}
\usepackage{booktabs}
\usepackage{makecell}
\usepackage{url}
\usepackage{cite}
\usepackage{amsmath}
\usepackage{color}
\usepackage{amssymb}
\usepackage{subfigure}
\usepackage{fancyhdr}
\usepackage{epic}
\usepackage{eepic}
\usepackage{threeparttable}
\usepackage{amscd}
\usepackage{here}
\usepackage{lscape}
\usepackage{tabularx}
\usepackage{longtable}
\usepackage{comment}
\usepackage{amsmath,amssymb,amsfonts}
\usepackage{textcomp}
\usepackage{float}
\usepackage{algorithm}
\usepackage{algorithmicx}
\usepackage{algpseudocode}
\usepackage{extarrows}
\usepackage{subcaption}
\definecolor{LightCyan}{rgb}{0.92,1,1}

\usepackage{lipsum}
\usepackage{tikz}
\definecolor{lime}{HTML}{A6CE39}
\DeclareRobustCommand{\orcidicon}{%
    \begin{tikzpicture}
        \draw[lime, fill=lime] (0,0)
        circle [radius=0.16]
        node[white] {{\fontfamily{qag}\selectfont \tiny ID}};
        \draw[white, fill=white] (-0.0625,0.095)
        circle [radius=0.007];
    \end{tikzpicture}
    \hspace{-2mm}
}

\foreach \x in {A, ..., Z}{%
    \expandafter\xdef\csname orcid\x\endcsname{\noexpand\href{https://orcid.org/\csname orcidauthor\x\endcsname}{\noexpand\orcidicon}}
}

\begin{document}

\title{From 0-to-1 to 1-to-N: Reproducible Engineering Evidence for MetaAI Recursive Self-Design}

\author{
Dun Li\orcidA{},
Jiatao Li,
Hongzhi Li

\thanks{
Dun Li is with the Department of Computing, The Hong Kong Polytechnic University, Hong Kong SAR, China. E-mail: dun.li@polyu.edu.hk
}

\thanks{
Jiatao Li is with the Merchant Marine College, Shanghai Maritime University, Shanghai 201306, China. E-mail: lijiatao@stu.shmtu.edu.cn
}

\thanks{
Hongzhi Li is with the College of Big Data and Artificial Intelligence, Chizhou University, Chizhou 247100, China. E-mail: pickpickup@sohu.com
}

\thanks{
Corresponding Author: Jiatao Li}
}

\markboth{IEEE Journal Draft,~Vol.~XX, No.~X, 2026}%
{Li \MakeLowercase{\textit{et al.}}: From 0-to-1 to 1-to-N}

\maketitle

\begin{abstract}
Recursive self-design refers to AI-assisted modification of the mechanisms by which an AI system is built, evaluated, and improved. This paper treats MetaAI not as a mature paradigm, but as a working term for a human-seeded, AI-expanded development pattern in which the design space itself becomes a target of modification. We propose an operational evidence framework with four criteria: inspectable target system, meta-level modifier, feedback-directed selection, and recursive continuation. We then map public systems, including Darwin G{\"o}del Machine (DGM), STOP, G{\"o}del Agent, and ShinkaEvolve, against these criteria. DGM provides the most direct currently reported evidence: its published results show improvement from 20\% to 50\% on SWE-bench Verified and from 14.2\% to 30.7\% on full Polyglot after 80 iterations, with ablations suggesting that both open-ended exploration and self-improvement contribute. Finally, we provide MetaAI-Mini, a reproducible HumanEval-based protocol and codebase. Because no completed model run is included in this build, MetaAI-Mini is reported as a protocol rather than as an experimental result.
\end{abstract}


\section{Introduction}

\IEEEPARstart{W}{e} use the term MetaAI to denote a human-seeded, AI-expanded development pattern in which human researchers specify an initial system, constraints, and evaluation protocol, while AI-driven processes propose and test successor designs. The term is used here as an analytical label rather than as a claim that a settled field or complete architecture already exists. The empirical question is therefore narrow: do public systems provide bounded evidence that AI can modify parts of the design space that determine later AI performance?

This idea has a long intellectual history. Good's ``ultraintelligent machine'' argument proposed that a sufficiently capable machine could design still better machines, producing a recursive amplification of intelligence \cite{good1965speculations}. Schmidhuber's G{\"o}del machine formalized self-modification as a proof-governed process in which a system rewrites itself only when it can prove the rewrite increases expected utility \cite{schmidhuber2003godel}. Later work discussed bounded recursive self-improvement and software-based seed AI, emphasizing both promise and risk \cite{nivel2013bounded,yampolskiy2015seed}. Although G{\"o}del's incompleteness theorem should not be interpreted as a direct limitation on AI architectures, the G{\"o}del-machine literature motivates the broader question of whether systems can modify their own improvement procedures under explicit criteria \cite{godel1931formal}.

Contemporary AI research has automated many design steps. AutoML, neural architecture search, and meta-learning can optimize model choices, architectures, and adaptation procedures inside human-specified spaces \cite{hutter2019automated,zoph2017neural,finn2017maml}. Large language model (LLM) agents extend fixed models with planning, memory, tools, and code execution \cite{wang2024agents}. Recent public systems also suggest that recursive self-improvement is becoming an organized research direction rather than a single isolated project: Sakana AI's RSI Lab explicitly frames DGM, ShinkaEvolve, and related systems as steps toward autonomous evolutionary optimization loops \cite{sakana2026rsilab}. Current methods predominantly perform optimization within a fixed, human-specified design space, whereas the pattern studied in this paper explicitly treats parts of that space as mutable and improvable objects.

The weakness of many discussions of recursive self-improvement is empirical. They either remain philosophical or jump directly to speculative superintelligence. This paper asks a narrower question: can we identify a real, public, reproducible experiment in which an AI system improves another AI system's code-level architecture or algorithmic scaffold, uses performance feedback to select improvements, and iterates this process across multiple generations? If so, such an experiment would not prove full AGI, but it would provide engineering evidence that recursive self-design can be operationalized in present-day systems.

We answer this question through a secondary empirical analysis of the Darwin G{\"o}del Machine (DGM) \cite{zhang2025dgm}. DGM is especially suitable because it uses a coding agent that modifies its own code repository, evaluates descendants on SWE-bench Verified and Polyglot, and maintains an archive of agents that can become parents for later self-modification. We also discuss STOP, G{\"o}del Agent, ADAS, SWE-agent, and ShinkaEvolve as related evidence, with citations given in the corresponding analysis below. The contribution is not a new DGM benchmark run; it is an evidence mapping over reported public systems plus a reproducible mini-protocol for future independent runs.

This paper makes three contributions:
\begin{itemize}
\setlength\itemsep{0em}\setlength\parsep{0em}\setlength\topsep{0em}
\item It defines four operational criteria for recursive self-design.
\item It maps DGM, STOP, G{\"o}del Agent, and ShinkaEvolve against those criteria, with ADAS and SWE-agent as adjacent public evidence.
\item It releases MetaAI-Mini, a reproducible HumanEval-based protocol and public GitHub codebase; because no API-backed run is included, only protocol configuration is reported.
\end{itemize}

The remainder of the paper is organized as follows. Section II defines operational criteria for recursive self-design. Section III describes the experimental design and the human 0-to-1 versus AI 1-to-N mapping. Section IV presents DGM results and visualizations. Section V introduces the MetaAI-Mini supplementary experiment. Section VI analyzes the strength and limits of the evidence, and Section VII concludes.

\section{Operational Criteria for Recursive Self-Design}

We first separate several terms that are often used interchangeably. \emph{Recursive self-design} denotes modification of the agent, scaffold, toolchain, prompt policy, evaluation workflow, or code-level mechanisms that shape future agent behavior. \emph{Recursive self-improvement} is broader and may include any iterative process in which a system improves its own future performance, including but not limited to design changes. \emph{Self-evolving agents} emphasizes lifelong adaptation across tasks and environments, but does not necessarily require the agent's own improvement procedure to be editable. \emph{LLM-driven program evolution} uses LLMs as mutation or search operators over programs and may evolve task solutions without making the agent scaffold itself the primary target.

We define recursive self-design as an iterative process satisfying four operational conditions:
\begin{enumerate}
    \item \emph{Inspectable target system}: there is a target AI system whose architecture, tools, workflow, prompts, memory, or code-level policy can be inspected and modified.
    \item \emph{Meta-level modifier}: an AI-driven process proposes changes to that target system, not merely random perturbations or human-authored patches.
    \item \emph{Feedback-directed selection}: proposed changes are evaluated on an external task or utility function, and the feedback influences which descendants are retained.
    \item \emph{Recursive continuation}: retained descendants can serve as the starting point for later rounds of improvement.
\end{enumerate}
Under this definition, full neural-weight self-rewriting is not required. A coding-agent scaffold can be a legitimate target if the scaffold materially changes how the underlying foundation model is used. This is important because present-day foundation models are often frozen, while much of agent performance depends on code-level tool use, context management, search, verification, and workflow design.

Table~\ref{tab:criteria_projects} summarizes how the main systems discussed in this paper align with the four operational conditions. DGM is the closest match because its descendants modify the code-level agent scaffold and can become later parents. STOP satisfies the conditions for the narrower object of an improver scaffold, G{\"o}del Agent satisfies them through runtime self-reference and task-driven policy modification, and ShinkaEvolve provides a useful program-evolution baseline where LLMs mutate scientific or algorithmic code under explicit evaluators.

\begin{table*}[t]
\centering
\caption{Operational criteria for recursive self-design across related systems}
\label{tab:criteria_projects}
\scriptsize
\setlength{\tabcolsep}{3pt}
\begin{tabular}{@{}p{0.10\textwidth}p{0.20\textwidth}p{0.19\textwidth}p{0.19\textwidth}p{0.20\textwidth}@{}}
\toprule
\textbf{System} & \textbf{Inspectable target system} & \textbf{Meta-level modifier} & \textbf{Feedback-directed selection} & \textbf{Recursive continuation} \\
\midrule
DGM & Full: repository, tools, prompts, and workflow are editable. & Full: archived agents propose and implement self-modifications. & Full: children are evaluated on SWE-bench or Polyglot before retention. & Full: retained descendants can become parents for later self-modification. \\
STOP & Partial: the improver scaffold is editable, but base LM weights are fixed. & Full: the seed improver is applied to its own code. & Full: meta-utility estimates downstream improvement quality. & Partial: continuation is bounded to scaffold-improvement rounds. \\
G{\"o}del Agent & Partial: runtime and policy-level agent logic are editable. & Partial: self-reference updates agent behavior, but not an archive of codebases. & Partial: task feedback guides policy changes across benchmarks. & Partial: recursive cycles are reported, without DGM-style lineage archives. \\
ShinkaEvolve & Limited: target programs are editable, not the full agent scaffold. & Partial: LLMs act as mutation operators over program candidates. & Full: fitness, novelty filtering, and adaptive sampling guide retention. & Not primary target: program populations evolve, while the improvement engine is not the main evolving object. \\
\bottomrule
\end{tabular}
\vspace{2pt}
\begin{minipage}{0.98\textwidth}
\footnotesize \emph{Note}: Full, Partial, Limited, and Not primary target are qualitative mappings based on the four operational criteria above. They summarize reported system designs and should not be read as independent benchmark replications. The labels indicate the degree to which each system satisfies the criterion for the specific target analyzed here, rather than an absolute ranking of system capability.
\end{minipage}
\end{table*}

\subsection{Distinction from Ordinary Optimization}

Recursive self-design is stricter than hyperparameter tuning. In ordinary optimization, a fixed search process adjusts parameters within a predefined space. In recursive self-design, the system may change the representation and procedure by which later tasks are solved. A temperature schedule or retry count alone would be weak evidence. A new editing primitive, a context summarization mechanism, or a patch-ranking workflow is stronger evidence because it changes the structure of the agent's interaction with tasks.

The distinction can be stated as follows. Boundary-internal optimization keeps the design space fixed:
\begin{equation}
D_{t+1}=D_t,\qquad x_{t+1}=\Omega(x_t;D_t).
\end{equation}
Recursive self-design changes the system that searches and acts:
\begin{equation}
S_{t+1}=\Psi(S_t,R_t,C_t),
\end{equation}
where $S_t$ is the current agent design, $R_t$ is empirical feedback, and $C_t$ is a constraint set such as sandboxing, benchmark protocols, and human oversight.

\section{Experimental Design}

\subsection{Experiment Selection}

The main experimental substrate is DGM because it most closely matches the four operational conditions above. It begins with one coding agent, lets archived agents self-modify their own codebases, evaluates the resulting agents on coding benchmarks, and uses an archive to preserve diverse stepping stones. The authors released code and experiment artifacts through a public repository \cite{dgmcode2025}. STOP and G{\"o}del Agent are included as secondary evidence because they report related forms of recursive scaffold improvement and self-referential agent update, but DGM is the primary experiment because it reports multi-iteration code-level agent evolution on standardized coding benchmarks. ShinkaEvolve is included as an external baseline and resource for sample-efficient program evolution rather than as the primary recursive self-design experiment: it evolves target programs using LLM mutation, novelty filtering, adaptive parent sampling, and bandit-based model selection, but the reported framework does not primarily make the full improvement engine itself the target of recursive modification.

\subsection{Benchmarks and Data}

DGM evaluates coding agents on two benchmarks. SWE-bench asks agents to resolve real-world GitHub issues by editing repositories and passing hidden tests \cite{jimenez2024swebench}. DGM uses SWE-bench Verified, a human-filtered subset intended to remove unsolvable tasks \cite{openai2024swebenchverified}. Polyglot evaluates coding ability across multiple programming languages and is associated with the Aider leaderboard \cite{gauthier2024polyglot}. The DGM report uses pass rates as the primary metric, i.e., the percentage of tasks solved by the agent.

The two benchmarks are useful for recursive self-design analysis because they test more than single-answer prediction. A coding agent must inspect files, edit code, manage context, run tests, and produce patches. Therefore, improvements to tools and workflow can change the agent's effective architecture even if the foundation model is frozen.

\subsection{Initial Target AI and Meta-AI Setting}

In DGM, the initial target AI is a coding agent built around a frozen foundation model with access to a Bash tool and an edit tool. For self-modification, the coding agents use Claude 3.5 Sonnet (New) in both the SWE-bench and Polyglot experiments. During benchmark evaluation, the SWE-bench agents use Claude 3.5 Sonnet (New), while Polyglot evaluation uses o3-mini \cite{anthropic2024claude35,openai2025o3mini}.

The meta-level improvement process is implemented as follows. The DGM archive stores generated agents. At each iteration, one or more parent agents are sampled from the archive. A parent analyzes its own benchmark logs, proposes a feature or workflow change, edits its own codebase to implement the proposal, and produces a child agent. The child is evaluated. If it compiles and retains basic code-editing functionality, it is added to the archive. The DGM experiments run for 80 iterations, generating one new agent per iteration, with parallelism in parent selection.

\begin{table}[t]
\centering
\fbox{\begin{minipage}{0.94\linewidth}
\small
\textbf{Protocol analyzed in this paper}\\
1. Human zero-to-one: define seed coding agent, tools, benchmarks, archive rule, sandbox, and evaluation protocol.\\
2. Initialize archive $A_0=\{S_0\}$.\\
3. For each iteration $t=1,\ldots,80$: sample parent(s) from archive; let parent inspect logs; let parent propose and implement a code-level self-modification; evaluate the child on staged benchmark tasks; retain child if it compiles and preserves code-editing functionality.\\
4. AI one-to-N: retained descendants become possible parents for future self-modification; the best archived agent is reported.
\end{minipage}}
\caption{Recursive self-design protocol instantiated by DGM. The outer archive and evaluation envelope are human-designed, while concrete agent modifications are generated by the AI-driven parent agents.}
\label{tab:protocol}
\end{table}

\subsection{Human Zero-to-One Versus AI One-to-N}

Table~\ref{tab:zeroone} maps the experiment to a human-seeded and AI-expanded division of labor. The human contribution is the seed and constraint envelope. The AI contribution is the repeated generation of concrete successor designs.

\begin{table}[t]
\centering
\caption{Mapping DGM to a human-seeded and AI-expanded division of labor}
\label{tab:zeroone}
\begin{tabular}{@{}p{0.24\linewidth}p{0.68\linewidth}@{}}
\toprule
\textbf{Role} & \textbf{Concrete instantiation in DGM} \\
\midrule
Human zero-to-one & Seed coding agent; initial Bash and edit tools; DGM archive procedure; parent-selection rule; staged evaluation protocol; benchmark choice; sandboxing and oversight. \\
AI one-to-N & Parent agents inspect evaluation logs, propose self-modifications, edit their own codebase, create child agents, and allow successful descendants to become future parents. \\
Fixed boundary & The open-ended exploration loop itself is not modified by DGM in the reported experiments. \\
Modified boundary & The agent's internal tools and workflows are modified, including editing primitives, context handling, and patch-generation procedures. \\
\bottomrule
\end{tabular}
\end{table}

\section{Evidence Mapping and Reported Results}

\subsection{Reported DGM Performance}

The primary external evidence is the endpoint improvement after 80 DGM iterations. We did not rerun DGM; Table~\ref{tab:mainresults} reports only values explicitly published in the DGM paper and repository. The first three rows are endpoint evaluations, while the last two rows add held-out cross-benchmark transfer checks reported by the DGM authors.

\begin{table}[t]
\centering
\caption{Published DGM endpoint and transfer results used in this secondary analysis; not newly rerun in this work}
\label{tab:mainresults}
\begin{tabular}{@{}lcc@{}}
\toprule
\textbf{Benchmark} & \textbf{Initial agent} & \textbf{Best DGM agent} \\
\midrule
SWE-bench Verified endpoint & 20\% & 50\% \\
Polyglot, 50-task staged endpoint & 14\% & 38\% \\
Polyglot, full-benchmark endpoint & 14.2\% & 30.7\% \\
SWE-bench $\rightarrow$ Polyglot transfer & 14.2\% & 28.9\% \\
Polyglot $\rightarrow$ SWE-bench transfer & 20\% & 24.5\% \\
\bottomrule
\end{tabular}
\vspace{2pt}
\begin{minipage}{0.96\linewidth}
\footnotesize \emph{Note}: The values are pass rates reported by the DGM authors. When the source reports a zero decimal digit, this table suppresses that trailing decimal for readability; nonzero decimals are preserved. The staged Polyglot setting uses 50 tasks, so each solved task changes the pass rate by 2 percentage points. The transfer rows evaluate the best agent evolved on one benchmark against the other benchmark.
\end{minipage}
\end{table}

These numbers support a bounded feasibility claim. A seed coding agent, without changing the underlying foundation model weights, generated descendants that solved substantially more tasks. The gain on SWE-bench Verified is 30 percentage points, and the full-Polyglot gain is 16.5 percentage points. The result is not a proof of unbounded self-improvement, but it is a concrete case in which recursive code-level self-design is associated with stronger agent performance.

\subsection{Reported DGM Ablation Evidence}

A stronger source of evidence is whether the recursive components matter in the published ablations. Table~\ref{tab:ablations} reports DGM's published comparison against ablations. Removing open-ended exploration sharply reduces performance. Removing self-improvement also reduces performance. A greedy parent-selection variant also underperforms the full archive-based DGM.

\begin{table}[t]
\centering
\caption{Published DGM ablation results used in this secondary analysis}
\label{tab:ablations}
\begin{tabular}{@{}lcc@{}}
\toprule
\textbf{Method} & \textbf{SWE-bench} & \textbf{Polyglot} \\
\midrule
DGM & 50\% & 38\% \\
DGM w/o open-ended exploration & 23\% & 14\% \\
DGM w/o self-improve & 39\% & 28\% \\
DGM greedy & 39.7\% & 30\% \\
\bottomrule
\end{tabular}
\end{table}

\begin{figure}[t]
\centering
\includegraphics[width=\columnwidth]{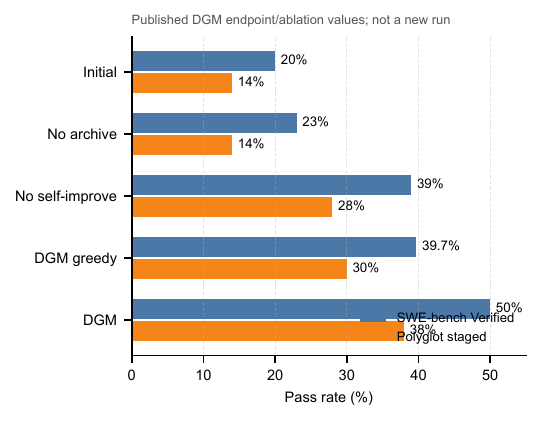}
\caption{Performance comparison using published DGM endpoint and ablation data. The Polyglot bars use the 50-task staged evaluation reported with the ablations; the full Polyglot endpoint is separately reported in Table~\ref{tab:mainresults}.}
\label{fig:bars}
\end{figure}

Figure~\ref{fig:bars} makes the ablation pattern visually explicit. If the full result came only from a fixed meta-agent repeatedly designing target agents, the ``DGM w/o self-improve'' baseline should be competitive. It is not. If the result came only from hill-climbing along the latest best lineage, the greedy or no-archive variants should be competitive. They are not. The reported evidence is therefore more consistent with a recursive, population-like process in which improved descendants and non-greedy stepping stones both contribute.

DGM also reports the percentage of generated SWE-bench agents that retain basic code-editing functionality: 51.3\% for full DGM, compared with 32.5\% for both the no-archive and no-self-improvement ablations. This matters because self-improvement requires preserving the ability to modify code. A system that improves task scores while destroying its capacity for further modification would not meet the recursive continuation criterion.

\subsection{Reported Structural Improvements}

Table~\ref{tab:structural} summarizes structural modifications reported in the DGM paper and repository. These are not merely scalar tuning operations. They change how the agent reads, edits, evaluates, and retries code.

\begin{table}[t]
\centering
\caption{Structural changes summarized from the DGM paper and repository; not newly discovered in this work}
\label{tab:structural}
\begin{tabular}{@{}p{0.34\linewidth}p{0.56\linewidth}@{}}
\toprule
\textbf{Change type} & \textbf{Reported effect on agent design} \\
\midrule
Granular file viewing & View relevant line ranges rather than entire files, reducing context burden. \\
String-replacement editing & Replace whole-file edits with precise substring replacement requiring unique matches. \\
Patch validation and retry & Detect empty or test-only patches and retry with source-file changes. \\
Context-window management & Summarize conversation history when the model context limit is reached. \\
Multiple patch generation and ranking & Generate several candidate patches and use another model or ranking procedure to select a stronger solution. \\
History-aware patch generation & Use previous attempts to condition later patch proposals. \\
\bottomrule
\end{tabular}
\end{table}

The DGM report further notes that similar high-level feature goals can lead to very different implementations and performance. For example, two SWE-bench nodes targeted finer-grained editing, but one implementation achieved 23.3\% while another achieved 40.5\%. This suggests an important point: recursive self-design is not only about choosing features; it is also about discovering implementation details that preserve agent functionality while improving downstream performance.

\subsection{Protocol Overview Without Synthetic Curves}

\begin{figure}[t]
\centering
\includegraphics[width=\columnwidth]{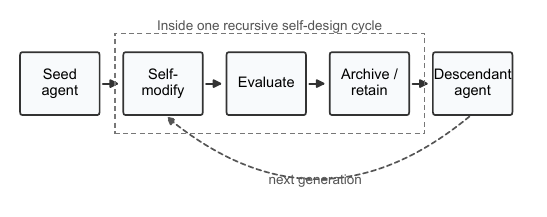}
\caption{MetaAI-Mini protocol overview. The diagram describes the experimental procedure---seed agent, self-modification, evaluation, archive/retention, and descendant agent---and is not an experimental performance measurement.}
\label{fig:protocol_overview}
\end{figure}

Figure~\ref{fig:protocol_overview} replaces a performance-trend schematic with a procedural overview. It shows the recursive loop used by MetaAI-Mini and related evidence mappings: a human-provided seed agent is modified, evaluated, either archived or rejected by a retention rule, and then used as the parent of a descendant generation. The figure is not a measurement and does not imply any unreported intermediate DGM or MetaAI-Mini performance values.

\subsection{Supporting Evidence from Related Systems}

STOP shows recursive scaffold improvement by applying a seed improver to its own code and selecting successors through a meta-utility \cite{zelikman2024stop}. ADAS directly studies automated design of agentic systems, where a meta-agent searches over code-level agent designs \cite{hu2024adas}. Both motivate DGM's stronger requirement that generated agents themselves remain capable of later self-modification.

G{\"o}del Agent reports self-referential policy updates across DROP, MGSM, MMLU, and GPQA, plus a 30-step MGSM improvement analysis \cite{yin2025godelagent}. SWE-agent is not a self-design system, but its agent-computer interface shows why coding-agent scaffolds and tool interfaces are performance-critical design objects \cite{yang2024sweagent}. Together they support the narrower claim that editable agent policies and interfaces can improve under feedback.

ShinkaEvolve is a complementary baseline for LLM-driven program evolution: LLM mutations, explicit evaluators, novelty filtering, adaptive parent sampling, and bandit model selection evolve target programs \cite{lange2025shinkaevolve,shinkaevolvecode2025}. It is not direct evidence of agent self-modification, but it is a useful comparison point for future MetaAI-Mini variants with deterministic evaluators.

\section{MetaAI-Mini: A Minimal Reproducible Experiment}

Full DGM reproduction is expensive because SWE-bench-style evaluation requires repository checkout, patch generation, hidden tests, and sandboxing. MetaAI-Mini is therefore included as a lightweight supplementary protocol, not as a benchmark result: a seed agent is evaluated, an LLM proposes a successor implementation, the candidate is tested, and the script retains it only if pass@1 improves.

The package is publicly available at \url{https://github.com/DunLi-Tsinghua/MetaAI-Mini} and uses the first 10 official OpenAI HumanEval records \cite{chen2021evaluating,humanevaldata2021}. It includes the official \texttt{HumanEval.jsonl.gz} file and a mechanically extracted \texttt{data/tasks.json}; it does not use 50 tasks or the full 164-task benchmark. The seed file \texttt{seed\_agent.py} exposes \texttt{solve\_task(task)}, while \texttt{self\_improve.py} logs generation, \texttt{run\_type}, active agent path, pass@1, and acceptance status. The same repository regenerates the PDF graphics inserted in this manuscript through \texttt{python analyze.py --paper-figures --skip-curve}. Generated Python code is executed locally with timeouts, so this is not a secure sandbox.

The mini-protocol is intentionally narrow but useful: it fixes the data split, records provenance checksums, and forces a distinction between smoke-test plumbing and API-backed improvement. This makes it suitable for classroom reproduction and for later replacement of the single-lineage rewrite loop with stronger search procedures.

No API-backed MetaAI-Mini run is reported in this manuscript. Mock smoke-test rows are marked \texttt{run\_type=mock} and must not be reported as results; real OpenAI-backed rows are marked \texttt{run\_type=api}. Table~\ref{tab:mini_results} is therefore a protocol-configuration table.

\begin{table}[t]
\centering
\caption{MetaAI-Mini protocol configuration}
\label{tab:mini_results}
\begin{tabular}{@{}p{0.32\linewidth}p{0.60\linewidth}@{}}
\toprule
\textbf{Field} & \textbf{Configuration} \\
\midrule
Dataset & Official OpenAI HumanEval downloaded from \texttt{HumanEval.jsonl.gz}. \\
Task count & First 10 official HumanEval records; not 50 tasks and not the full 164-task benchmark. \\
Seed agent & \texttt{seed\_agent.py}, exposing \texttt{solve\_task(task)}. \\
Model & Configurable through \texttt{OPENAI\_MODEL}; default script value is \texttt{gpt-4.1-mini}. \\
Generations & Default 5 candidate-improvement generations. \\
Evaluation & Local subprocess execution with per-task timeout; measured pass@1 on the 10-task subset. \\
Retention rule & Retain a candidate only if its pass@1 strictly improves over the active agent. \\
Current status & Public GitHub package released; protocol checked on real downloaded data; no MetaAI-Mini performance values are reported without \texttt{run\_type=api} rows. \\
\bottomrule
\end{tabular}
\end{table}

The analysis script generates a formal improvement curve only from API-backed rows by default; \texttt{--allow-mock} is reserved for smoke-test plotting. This keeps the manuscript aligned with completed runs rather than intended or mock runs.

\section{Analysis}

The evidence is consistent with a limited MetaAI-style pattern. In DGM, humans define the seed, benchmark, safety boundary, and archive process; the AI modifies code-level agent design, and successful descendants become future parents. Figure~\ref{fig:bars} shows that the full DGM loop outperforms ablations, while Figure~\ref{fig:protocol_overview} separates protocol structure from measurement. STOP and G{\"o}del Agent provide adjacent support through recursive improver updates and self-referential policy modification; ShinkaEvolve clarifies what evaluator-driven program evolution can achieve when the evolving object is a target program rather than the agent scaffold.

The most important inference is structural rather than merely quantitative. DGM's gains arise from changes to the agent's editing tools, retry logic, context handling, and candidate-ranking workflow. These are not final AGI mechanisms, but they are concrete examples of an AI system modifying the operational substrate through which future task attempts are made.

The archive appears central. A greedy process may discard low-scoring variants that later become useful stepping stones, whereas DGM's archive-based process preserves multiple lineages. This suggests that future MetaAI-style systems should compare single-chain self-edits with evolutionary archives, lineage audits, rollback mechanisms, and diversity-preserving parent selection.

The evidence is bounded. It does not establish full RSI or AGI: the foundation models remain fixed, the benchmark suite and retention rules are human-authored, and DGM does not modify its outer open-ended exploration loop. The experiments are also expensive, stochastic, and coding-centric. The results should therefore be read as feasibility evidence for recursive self-design, not as evidence for an imminent intelligence explosion. A stronger next step would report machine-readable lineage data, per-generation diffs, evaluation traces, and failed modifications so that self-design dynamics can be audited rather than inferred from endpoint scores alone.

Safety is integral rather than secondary. A self-modifying system can alter tools, evaluators, logging, context handling, and execution pathways. The cited STOP and DGM artifacts both indicate that future systems need sandboxing, tripwires, independent evaluation, immutable audit logs, and human-governed deployment gates.

\section{Conclusion}

This paper proposed an operational evidence framework for recursive self-design and applied it to public systems. DGM is the clearest reported example we identify: after 80 iterations, a seed coding agent improves from 20\% to 50\% on SWE-bench Verified and from 14.2\% to 30.7\% on full Polyglot, with ablations indicating that self-improvement and open-ended exploration both matter. Together with STOP, G{\"o}del Agent, ShinkaEvolve, and MetaAI-Mini, these systems provide bounded evidence that recursive self-design and adjacent LLM-driven program evolution are plausible engineering patterns.

The contribution is deliberately limited: the paper does not claim autonomous design of arbitrary future intelligence and reports no MetaAI-Mini performance without an API-backed run. Future work should publish per-iteration logs, test beyond coding, make outer improvement loops modifiable under safeguards, and evaluate safety as a first-class outcome.

\bibliographystyle{IEEEtran}
\bibliography{references}

\end{document}